\documentclass[pmlr,twocolumn,10pt]{jmlr} 

\usepackage{graphicx}
\usepackage{tikz}
\usepackage{amsmath,amssymb,amsfonts} 

\usepackage{booktabs}
\usepackage{colortbl}
\usepackage{makecell}
\usepackage{multirow}
\usepackage[table,dvipsnames]{xcolor}
\usepackage{enumitem}

\usepackage{caption}
\usepackage{subcaption}
\usepackage{titlesec}
\titlespacing\section{0pt}{3pt plus 2pt minus 0pt}{3pt plus 2pt minus 0pt}
\titlespacing\subsection{0pt}{2pt plus 2pt minus 0pt}{2pt plus 2pt minus 0pt}

\usepackage{pifont}

\usepackage{siunitx}

\usepackage{calc}

\definecolor{matplotlib_blue}{HTML}{1F77B4}
\definecolor{matplotlib_orange}{HTML}{FF7F0E}
\definecolor{matplotlib_green}{HTML}{2CA02C}
\definecolor{matplotlib_red}{HTML}{D62728}
\definecolor{bright_gray}{HTML}{EEEEEE}

\newcommand{\KPSC}{Proprietary}

\renewcommand{\dagger}{\textsuperscript{\textdagger}}

\DeclareMathOperator*{\argmin}{argmin}

\DeclareMathOperator*{\mean}{mean}
\newcommand{\operation}[1]{\operatorname{#1}}

\newcommand{\embedding}[3]{
    \draw[fill=bright_gray] (#1,#2) rectangle (#1+1,#2+0.25);
    \draw[fill=#3] (#1+0.125,#2+0.125) circle (0.125);
    \draw[fill=#3] (#1+0.375,#2+0.125) circle (0.125);
    \draw[fill=#3] (#1+0.625,#2+0.125) circle (0.125);
    \draw[fill=#3] (#1+0.865,#2+0.125) circle (0.125);
}

\newcommand{\equal}[1]{{\hypersetup{linkcolor=black}\thanks{#1}}}

\theorembodyfont{\upshape}
\theoremheaderfont{\scshape}
\theorempostheader{:}
\theoremsep{\newline}

\jmlrvolume{333}
\jmlryear{2026}
\jmlrsubmitted{LEAVE UNSET}
\jmlrpublished{LEAVE UNSET}
\jmlrworkshop{Conference on Health, Inference, and Learning (CHIL) 2026} 

\title[A Multi-Dataset Benchmark of Multiple Instance Learning for 3D Neuroimage Classification]{A Multi-Dataset Benchmark of Multiple Instance Learning\\for 3D Neuroimage Classification}

\author{%
    \Name{Ethan Harvey}$^1$\equal{Equal contribution.} \Email{ethan.harvey@tufts.edu} \\
    \Name{Dennis Johan Loevlie}$^1$\footnotemark[1] \Email{dennis.loevlie@tufts.edu} \\
    \Name{Amir Ali Satani}$^2$ \Email{amir.satani@tufts.edu} \\
    \Name{Wansu Chen}$^3$ \Email{wansu.chen@kp.org} \\
    \Name{David M. Kent}$^2$ \Email{david.kent@tuftsmedicine.org} \\
    \Name{Michael C. Hughes}$^1$ \Email{michael.hughes@tufts.edu} \\
    \addr $^1$Department of Computer Science, Tufts University, Medford, MA, USA \\
    \addr$^2$Predictive Analytics and Comparative Effectiveness Center, Tufts Medical Center, Boston, MA, USA \\
    \addr$^3$Department of Research and Evaluation, Kaiser Permanente Southern California, Pasadena, CA, USA \\
}

\begin{document}

\maketitle

\begin{abstract}
    Despite being resource-intensive to train, 3D convolutional neural networks (CNNs) have been the standard approach to classify CT and MRI scans.
    Recent work suggests that deep \emph{multiple instance learning} (MIL) may be a more efficient alternative for 3D brain scans, especially when the pre-trained image encoder used to embed each 2D slice is frozen and only the pooling operation and classifier are trained.
    In this paper, we provide a systematic comparison of simple MIL, attention-based MIL, 3D CNNs, and 3D ViTs across three CT and four MRI datasets, including two large datasets of at least 10,000 scans.
    Our goal is to help resource-constrained practitioners understand which neural networks work well for 3D neuroimages and why.
    We further compare design choices for attention-based MIL, including different encoders, pooling operations, and architectural orderings.
    We find that simple mean pooling MIL, without any learnable attention, matches or outperforms recent MIL or 3D CNN alternatives on 4 of 6 moderate-sized tasks. This baseline remains competitive on two large datasets while being 25x faster to train.
    To explain mean pooling's success, we examine per-slice attention quality and a semi-synthetic dataset where we can derive the best possible classifier via a Bayes estimator. This analysis reveals the limits of existing MIL approaches and suggests routes for future improvements.
\end{abstract}

\begin{figure}[!t]
    \centering
    \includegraphics[width=\linewidth]{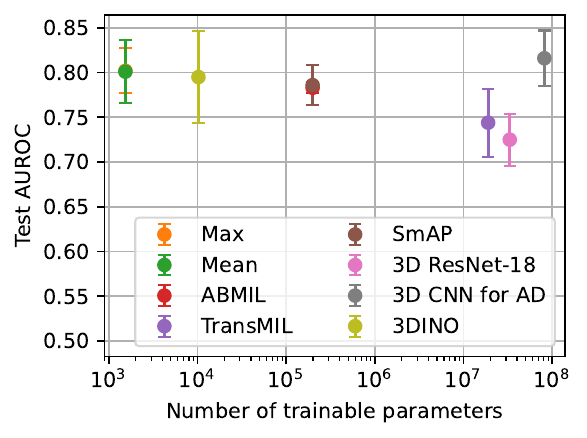}
    \caption{Test AUROC on OASIS-3 MRI (higher is better) vs. number of trainable parameters (lower is better) for MIL and 3D CNN methods.
    \textbf{Takeaway: {\color{matplotlib_green} Mean pooling MIL} is almost as good and far more efficient than the best 3D CNN or MIL method.} All MIL methods use the same pre-trained ViT encoder.
    }
    \label{fig:oasis-3_mri_number_of_trainable_parameters}
\end{figure}

\paragraph*{Data and Code Availability.}
We use open-access datasets: ADNI1~\citep{mueller2005ways}, OASIS-3~\citep{lamontagne2019oasis}, and RSNA~\citep{flanders2020construction}, as well as non-shareable \KPSC\ data (Sec.~\ref{sec:experiments}). Our code is publicly available: \href{https://github.com/tufts-ml/neuroimage-classifiers}{\texttt{github.com/tufts-ml/neuroimage-classifiers}}.

\paragraph*{Institutional Review Board (IRB).} 
Our study of Proprietary datasets of deidentified neuroimages with associated diagnostic labels was approved by our local Institutional Review
Board (Tufts Health Science IRB \#3977 and \#4374). Other datasets are open-access and deidentified, so no approval is needed.


\section{Introduction}
\label{sec:introduction}

We investigate the problem of predicting a single binary label given a 3D image of the brain. We focus on two common imaging modalities: computed tomography (CT) and magnetic resonance imaging (MRI).
Such images provide high-quality clinical evidence for critical neurological conditions such as Alzheimer's or brain lesions.
Our work is especially motivated by the potential for effective detection of small vessel diseases like covert brain infarction or white matter disease. Reliable automatic detection on routine scans may improve long-term risk assessment for stroke, dementia, and other serious consequences~\citep{pasi2020clinical}.

Deep neural networks have become widespread for neuroimaging classification tasks.
One approach allows the network direct access to an input 3D volume~\citep{wen2020convolutional}.
Another increasingly common approach is \emph{multiple instance learning} (MIL).
Here, the 3D volume is divided into slices.
Each slice or ``instance'' is processed separately by an encoder, then the MIL architecture pools instance-level results into a whole-scan prediction.
We consider here both simplistic pooling strategies as well as recent pooling methods like TransMIL~\citep{shao2021transmil} and SmAP~\citep{castro2024sm} that account for interaction between instances.

Two aspects of MIL make it appealing for neuroimage classification.
First, current MIL processing broadly parallels how human clinical experts make classification judgments from 3D images. Experts today page through axial slices to find a specific slice or set of slices with visual evidence for an abnormality, synthesizing information across slices with clinical context.
The ability to find one ``smoking gun'' instance to justify a positive classification of the entire scan is a fundamental assumption of MIL \citep{dietterich1997solving,raff2023reproducibility}.
Second, MIL neural networks can be more \emph{runtime and storage efficient} for training and evaluation than 3D neural networks, especially if using a pre-trained encoder. 

We make several contributions to improve the research community's understanding of MIL strengths and weaknesses for 3D brain scan classification:
\begin{itemize}[leftmargin=*,noitemsep,topsep=0pt]
    \item We provide a systematic study of MIL architectural approaches, encoders, and pooling operations on 9 classification tasks using 7 datasets of MRI or CT scans (Tab.~\ref{tab:dataset_distributions}). This multi-dataset, apples-to-apples comparison is missing in earlier work trying MIL for brain imaging~\citep{castro2024sm}, which compare advanced MIL methods but not 3D neural networks or simple baselines like mean pooling on one relatively small CT dataset.  
    \item We provide evidence across multiple MRI and CT datasets that a simple MIL baseline using mean pooling, rather than learned attention, can often match or outperform recent MIL methods published in top conferences, despite the substantial additional complexity of learned attention. Even on larger datasets with 10,000+ scans, we find recent variants of attention-based MIL like TransMIL~\citep{shao2021transmil} or SmAP~\citep{castro2024sm} never provide an absolute gain in AUROC greater than 0.025~(see Tab.~\ref{tab:large_datasets_embedding-aggregation_auroc}), while requiring 25x longer training times. We thus argue for including mean pooling MIL as a strong baseline in future work.
    \item We directly examine the quality of learned attention on real data.
    The RSNA CT dataset~\citep{flanders2020construction} uniquely provides instance-level (per-slice) labels for lesion presence/absence, enabling evaluation of how well learned attention predicts instance-level labels. We compare ABMIL, TransMIL, and SmAP to a simple center-focused Gaussian baseline that \emph{ignores the image entirely} and allocates attention solely by slice position. Surprisingly, on this large dataset no learned attention method outperforms this trivial baseline on attention correctness, AUROC, or AUPRC (see Tab.~\ref{tab:rsna_instance-level_metrics}).
    \item To better explain the lack of strong gains from recent MIL methods designed to account for instance interaction, in Sec.~\ref{sec:semi_synth} we create a semi-synthetic dataset intended to match the statistics of slice-level labels in the RSNA CT dataset. Knowledge of the data-generating process allows us to define the best possible classifier via a Bayes estimator.
    This analysis reveals that even at large data sizes, recent MIL approaches score substantially worse than the best possible classifier.
\end{itemize}
Altogether, our work suggests the potential for future methods innovation in the MIL design space and provides a reproducible platform for verifying how such innovations impact overall classifier quality.

\section{Background: MIL Methods}
\label{sec:background}

MIL~\citep{dietterich1997solving,maron1997framework,quellec2017multiple} is a branch of weakly supervised learning where the goal is to train a predictor that, given a variable-sized set of instances each with its own feature vector, can predict a single binary label for the entire set.
The training dataset for a generic MIL problem, denoted $\{(x_i, y_i)\}_{i=1}^N$, consists of $N$ labeled bags of data. Each bag is a set of $S_i$ instance feature vectors $x_i = \{x_{i,1}, \ldots, x_{i,S_i}\}$ with a single binary label $y_i \in \{0, 1\}$.
For our work on brain scans, each bag is a 3D CT or MRI, and an instance is the 2D image of one axial slice.

In this work, we study three aspects of MIL design for 3D brain scans: architectural approach, encoder choice, and pooling operation. These are summarized in Tab.~\ref{tab:mil_components} and outlined in the three subsections below.

\setlength{\tabcolsep}{2pt}
\begin{table}[!h]
    \centering
    \caption{We provide a systematic comparison of MIL architectural approaches, encoders, and pooling operations.}
    \label{tab:mil_components}
    \small
    \resizebox{\linewidth}{!}{\begin{tabular}{ll}
        \hline
        \textbf{Architectural ordering} & \makecell[lt]{\emph{Embedding-aggregation},\\\emph{prediction-aggregation}} \\
        \hline
        \textbf{Encoders} & \makecell[lt]{ViT-B/16, ConvNeXt-Tiny,\\MedSAM} \\
        \hline
        \multicolumn{2}{l}{\textbf{Pooling}} \\
        \textit{~~MIL without instance interaction} & Max, Mean, ABMIL \\
        \textit{~~MIL with instance interaction} & TransMIL, SmAP \\
        \hline
    \end{tabular}}
\end{table}
\setlength{\tabcolsep}{6pt}

\subsection{Architectural ordering approaches}
Among deep neural networks for MIL, 
there are two main paradigms, which we refer to as \emph{embedding-aggregation} and \emph{prediction-aggregation}.
Both approaches consist of three parts: an encoder, a pooling operation, and a classifier.
They differ in the ordering of these parts, as shown in  Fig.~\ref{fig:architectural_approaches}. 

\input{fig_architectural_approaches}

In the \emph{embedding-aggregation} approach, the order
is encode, pool, then classify.
First, each instance's $B$-channel 2D image $x_{i,j} \in \mathbb{R}^{B \times H \times W}$ is encoded to an instance-specific representation vector $h_{i,j} = f(x_{i,j}) \in \mathbb{R}^M$.
Second, a pooling operation $\sigma$ (e.g., max, mean, or attention-based pooling) aggregates all $S_i$ instance representations $h_i = \{h_{i,1}, \ldots, h_{i,S_i}\}$ into a single representation vector $z_i = \sigma(h_i) \in \mathbb{R}^M$.
Finally, the bag level representation vector $z_i$ is classified into a predicted probability vector over $C$ classes, $g(z_i) \in \Delta^C \subset \mathbb{R}^C$. We can denote the ultimate prediction as $\hat{y}_i = g(\sigma(f(x_i))$. In this notation, applying $f$ to a set yields another set containing a mapping of each instance.

In the prediction-aggregation approach, the ordering of $g(\cdot)$ and $\sigma(\cdot)$ is swapped.
A separate prediction score vector containing logits or probabilities is produced for each of the $S_i$ instances separately, and then pooling determines the final prediction, $\hat{y}_i = \sigma(g(f(x_i)))))$.

In either approach, model parameters for all parts (encoder, pooling, and classifier) can be trained to minimize binary cross entropy averaged across all data: $\frac{1}{N} \sum_{i=1}^N \ell^{\text{CE}}(y_i, \hat{y}_i)$, where $\hat{y}_i$ is a function of input features and parameters. 

\subsection{Encoders}
In both architectural approaches, an encoder $f(\cdot)$ embeds each 2D slice into a representation vector.
We compare a vision transformer (ViT, \citealt{dosovitskiy2021image}) pre-trained on ImageNet-1k \citep{deng2009imagenet}, a convolutional encoder \citep{liu2022convnet} also pre-trained on ImageNet-1k, and a medical image encoder from the medical segment anything model (MedSAM, \citealt{ma2024segment}), a variant of a recent method~\citep{kirillov2023segment} fine-tuned for medical image segmentation on a large medical image dataset.

\subsection{Pooling}
The design of the pooling operation $\sigma(\cdot)$, which aggregates across instances, is generally most important for understanding how spatial context is incorporated. We describe several architectures below. We focus on \emph{embedding-aggregation} for concreteness; translation to \emph{prediction-aggregation} is straightforward.
Here, we take as input a set of embeddings $h_i = \{h_{i,1}, \ldots, h_{i,S_i}\}$ for bag $i$.
Each instance $j$ in the bag is encoded as a representation vector $h_{i,j} \in \mathbb{R}^M$.

\textbf{Max and Mean pooling.} Two simple pooling operations find the maximum or mean \emph{\textbf{element-wise}} of the given $M$-dimensional vectors:
\begin{align}
    z_{i} = \max_{j=1,\dots,S_i} h_{ij}, \quad \text{or} \quad
    z_{i} = \mean_{j=1,\dots,S_i} h_{ij}.
\end{align}
Early deep MIL methods~\citep{pinheiro2015image,zhu2017deep,feng2017deep} used such simple, non-trainable operations aggregate instance representations.

\textbf{Attention-based pooling.}
Attention-based pooling~(ABMIL)~\citep{ilse2018attention} assigns an attention weight $a_{ij}$ to each instance via
\begin{align}
    a_{ij} = \frac{\exp\left(u^\top \tanh\left(U h_{ij}\right)\right)}{\sum_{k=1}^{S_i} \exp\left(u^\top \tanh\left(U h_{ik}\right)\right)},
    \label{eq:abmil}
\end{align}
then forms bag-level embedding vector $z_i$ via a weighted average: $z_i = \sum_{j=1}^{S_i} a_{ij} h_{ij}$. The weights $a_{ij}$ are non-negative and sum to one: $a_{ij} \ge 0$ for all $j$; $\sum_j a_{ij} = 1$. 
In the equation above, vector $u \in \mathbb{R}^L$ and matrix $U \in \mathbb{R}^{L \times M}$ are trainable parameters.

\textbf{Smooth attention pooling.}
Smooth attention pooling (SmAP) \citep{castro2024sm} uses a smoothing operation to add local interactions between instance embeddings.
The smoothed embeddings $g_i \in \mathbb{R}^{S_i \times M}$ for all $S_i$ instances are obtained by solving an optimization problem
\begin{align}
    \operation{\texttt{Sm}}(h_i) &= \argmin_{g_i} \alpha\mathcal{E}_D(g_i) + (1-\alpha) \|h_i - g_i\|^2_F,
\end{align}
where $\alpha \in [0, 1)$ controls the amount of smoothness, $\| \cdot \|_F$ denotes the Frobenius norm, and
\begin{align}
\textstyle    \mathcal{E}_D(g_i) &= \textstyle \frac{1}{2} \sum_{j=1}^{S_i} \sum_{k=1}^{S_i} A_{ijk} \|g_{ij} - g_{ik}\|^2_2.
\end{align}
Here, $A_i \in \mathbb{R}^{S_i \times S_i}$ is an adjacency matrix defining local relationships between instances and $\| \cdot \|^2_2$ denotes the squared Euclidean norm aka ``sum of squares''.
App.~\ref{sec:smap_varying_number_of_neighbors} has a detailed investigation of ways to set $A$ for 3D neuroimages; results in the main paper just link each slice to its adjacent slices in scan order, as in ~\citep{castro2024sm}.
The resulting smoothed embedding $g_{ij}$ then replaces embedding $h_{ij}$ in Eq.~\eqref{eq:abmil}.

\textbf{Transformer-based pooling.}
\citet{shao2021transmil}'s 
transformer-based correlated MIL (TransMIL) allows instance interactions to inform pooling.
First, TransMIL uses convolutions over instances in a pyramidal position encoding to model dependencies. Second, interactions between \emph{all pairs} of instances are captured via multi-head self-attention. For layer $\ell$ and head $h$, there's a
$S_i {+} 1 \times S_i {+} 1$ attention matrix, where rows sum to one and weight $j,k$ is:
\begin{align}
    a_{i,j,k}^{(\ell,h)} \propto \exp \left( \left(q_{i,j}^{(\ell,h)}\right)^\top k_{i,k}^{(\ell,h)} / \sqrt{D} \right).
\end{align}
Here, each instance $j$ has embeddings of size $D$ for query $q_{i,j}^{(\ell,h)} = W_Q^{(\ell,h)} h_{i,j}^{\ell-1}$, key $k_{i,j}^{(\ell,h)} = W_K^{(\ell,h)} h_{i,j}^{\ell-1}$, and value $v_{i,j}^{(\ell,h)} = W_V^{(\ell,h)} h_{i,j}^{\ell-1}$.
Propagating embeddings via attention-weighted value averages over several layers and heads allows instance features to interact flexibly to inform the ultimate bag-level embedding.

\section{Related Work}
\label{sec:related_work}

\textbf{CNNs for 3D neuroimaging.} 
Deep learning has been widely applied to neuroimaging classification tasks \citep{huang2023self,dorfner2025review}.
Several works examine 2D vs. 3D CNN architectures.
\citet{wen2020convolutional} provide a comprehensive overview and reproducible evaluation of CNNs for Alzheimer's disease (AD) classification using MRI neuroimages.
They compared 2D slice-based and 3D volumetric approaches on the ADNI, AIBL and OASIS datasets and found that the performance of 2D slice-based approaches pre-trained on ImageNet-1k \citep{deng2009imagenet} was generally lower compared to 3D approaches.
On a small subset, \citet{dufumier2021benchmarking} showed that randomly initialized 3D CNNs outperform randomly initialized 2D CNNs using mean pooling with a \emph{prediction-aggregation} architecture.
Recent work further demonstrates that the performance of 3D CNNs strongly depends on design choices related to normalization, downsampling, and depth~\citep{liu2019design,liu2022generalizable}.
In contrast to these works that focus only on 2D and 3D CNN approaches, we systematically compare MIL methods.

\textbf{MIL for neuroimaging.}
The popularity of MIL for neuroimaging classification tasks has grown in recent years \citep{tong2014multiple,lopez2022deep,harvey2023probabilistic,perez2024end}, yet quality benchmarking especially to methods outside MIL remains underexplored. For example, 
\citet{wu2021combining} in their Table 2 compare their MIL method's ICH classification results on a subset of the RSNA dataset to 3D CNN ICH classification numbers pulled directly from other papers that evaluate on \emph{different datasets}. 
This comparison is ``apples-to-oranges'', making it difficult to draw conclusions about the relative rankings of the two approaches.

Recent MIL for neuroimaging work has focused on improving classification and localization results compared to conventional MIL by introducing smooth attention~\citep{wu2023smooth,castro2024sm}, described earlier in Sec.~\ref{sec:background}.
However, despite growing interest, there are no systematic evaluations of different MIL encoders, aggregation approaches, and pooling operations.
Moreover, MIL's relative strengths and weaknesses compared to more traditional approaches like 3D CNNs remain poorly understood.

\textbf{Brain foundation models.}
Work has begun to develop brain foundation models that pre-train large encoders on neuroimaging data and then adapt them to downstream neuroimaging tasks.
For example, \citet{deng2026brain} propose a brain foundation model that they adapt for brain disease segmentation and classification tasks, and \citet{wang2024enhancing} introduce Vote-MI, an unsupervised representative slice selection method that selects representative 2D slices from 3D brain MRIs to enable effective use of 2D vision–language models (VLMs).
Related multimodal clinical foundation models, such as \citet{dai2025qoqmed}, also include 3D neuroimages in their training data \citep{dai2025climb}.
Overall, these efforts are relevant but largely tangential to our study. 
Many brain foundation model pipelines ultimately process 3D neuroimages in a slice-wise manner, which requires aggregating information across slices to produce scan-level predictions. 
Therefore, our systematic evaluation of different MIL architectural orderings and pooling operators can inform the design of future MIL that uses a brain foundation model encoder.

\section{Experiment Design}
\label{sec:experiments}

\subsection{Datasets of 3D Brain Scans}

We study classification on the seven datasets of 3D brain scans in Tab.~\ref{tab:dataset_distributions}.
For all CT datasets, we follow recommended preprocessing~\citep{muschelli2019recommendations}, leaving the original number of slices which varies across scans.
For all MRI datasets, we follow recommended preprocessing~\citep{wen2020convolutional,routier2021clinica}. Each available modality (T1 and T2, if included) is mapped to a fixed-size template with 179 slices.  
Across CT and MRI, the same preprocessed 3D image is fed into all 3D and MIL NNs.
MIL encoders process axial slices where each 2D image is resized to $224 \times 224$ (except for MedSAM, which uses $1024 \times 1024$).
See App.~\ref{sec:preprocessing} for more details preprocessing.

\setlength{\tabcolsep}{6pt}
\begin{table*}[htbp!]
    \centering
    \caption{Dataset statistics: Patient-scan counts by class.}
    \label{tab:dataset_distributions}
    \small
    \begin{tabular}{lllrrrr}
        \hline
        \textbf{Dataset} & \textbf{Modality} & \textbf{Label} & \textbf{Num Neg.} & \textbf{Num Pos.} & \textbf{Total Scans} 
        & \textbf{Instances/Scan}
        \\
        \hline
        \rowcolor{bright_gray} ADNI1 & MRI & AD & 1,818 & 476 & 2,294 & 179-179 \\
        OASIS-3 & CT & AD & 556 & 106 & 662 & 74-111 \\
        & MRI & AD & 1,239 & 381 & 1,620 & 179-179 \\
        \rowcolor{bright_gray} RSNA-1,149 & CT & ICH & 666 & 483 & 1,149 & 24-57 \\
        RSNA-21,744 & CT & ICH & 12,862 & 8,882 & 21,744 & 20-60 \\
        \rowcolor{bright_gray} \KPSC-800 & MRI & CBI & 549 & 251 & 800 & 179-179 \\
        \rowcolor{bright_gray} & & WMD & 259 & 541 & 800 & 179-179 \\
        \KPSC-10k & MRI & CBI & 9,526 & 474 & 10,000 & 179-179 \\
        & & WMD & 5,709 & 4,291 & 10,000 & 179-179 \\
        \hline
    \end{tabular}
\end{table*}
\setlength{\tabcolsep}{6pt}

\textbf{ADNI1.}
The ADNI1 Complete 1Y 1.5T dataset~\citep{mueller2005ways} includes 2,294 T1 MRI scans from 639 patients.
We use the diagnostic cohorts assigned upon enrollment for binary classification of Alzheimer's disease (AD).

\textbf{OASIS-3 CT.}
The OASIS-3 CT dataset~\citep{lamontagne2019oasis} has 662 CT scans from 495 patients.
We use the clinical dementia rating (CDR) for binary classification of AD.
Positive labels indicate the patient has a diagnosis at most 80 days before or 365 days after the MRI scan date.

\textbf{OASIS-3 MRI.}
The OASIS-3 MRI dataset~\citep{lamontagne2019oasis} includes 1,620 T1 and T2 MRI scans from 903 patients.
Label definitions for AD are the same as in OASIS-3 CT.

\textbf{RSNA-1,149.}
Prior work in MIL for neuroimage classification~\citep{wu2021combining,lopez2022deep,perez2024end,castro2024sm} uses a subset of 1,150 CT scans from the RSNA 2019 Brain CT Hemorrhage Challenge~\citep{flanders2020construction}.
In the released subset's train and test sets \citep{castro2025rsna}, we found one scan appears in both the training and test set.
We removed the duplicate scan, resulting in 1,149 unique CT scans which we subdivide as described below.

\textbf{RSNA-21,744.}
The  RSNA 2019 Brain CT Hemorrhage Challenge~\citep{flanders2020construction} includes 752,803 slices from 21,744 CT scans with released labels for presence/absence of any intracranial hemorrhage (ICH).
We subdivide the released training set (no other release has labels) into our own training, validation, and test sets as described below.

\textbf{\KPSC-800.}
The \KPSC-800 dataset contains de-identified axial T1 and T2 MRI scans from 800 patients 50+ years of age who received an MRI in 2009-2019 during the course of routine care within the Kaiser Permanente health system in southern California, USA. 
Our study's IRB approval is documented on page 1.

Two separate binary labels are of interest: white matter disease (WMD) and covert brain infarction (CBI). These are often incidental findings with no outward symptoms. Yet the presence of either may be predictive of future stroke or dementia~\citep{pasi2020clinical}.

Of the 800 scans here, 640 scans were randomly sampled from all eligible scans. 
For the remainder, we deliberately sample 160 scans directly from CBI-positive patients in the eligible cohort (regardless of WMD status) to overcome CBI's rarity.
Our 800 scan dataset contains 251 CBI cases and 541 WMD cases.

\textbf{\KPSC-10k.}
The \KPSC-10k dataset is a larger dataset of T1 and T2 MRIs from 10,000 patients 50+ years of age who received a routine-care MRI in 2009-2019 from the same southern California health system. 
Our study's IRB approval is documented on page 1.

This larger dataset has the same binary labels for WMD and CBI.
All scans with CBI (regardless of WMD status) were included and the remaining scans were randomly sampled from the eligible cohort.
This yielded a total dataset of 474 CBI cases and 4,291 WMD cases.

\textbf{WMD/CBI label extraction.}
WMD and CBI labels for Proprietary scans are extracted from routine text reports provided by an clinical expert during routine image interpretation.
We use an NLP tool developed by~\citet{fu2019natural} for the extraction.

\textbf{Spatial extent of labels.} The relevant spatial extent of a 3D scan needed to correctly classify the different labels here differs in important ways that impact our later analysis.
The binary label for AD arises from broader clinical knowledge not just observed imaging. In imaging, the signs of neurodegenerative diseases like AD are likely diffuse throughout many axial slices.
In contrast, ICH lesions are visible in a focal region of the brain but can be volumetrically extensive: in RSNA the mean number of contiguous slices for a lesion is 12.
CBI lesions are also focal, lacunar infarcts in particular are often less than 1 cm in size.
WMD is typically bilateral and regionally patterned (periventricular and deep white matter), often visible across multiple slices. It is regionally concentrated, but not as focal as a lacunar infarct.

\subsection{3D CNN Methods}

3D CNNs are the standard approach to classify CT and MRI scans.
We include a 3D ResNet-18 that can process a variable number of inputs \citep{tran2018closer} and a 3D CNN designed for AD \citep{liu2019design,liu2022generalizable} as baselines.

\subsection{Training and Evaluation Procedures}

For all non-\KPSC~datasets, we randomly assign images at a 4:1:1 ratio into training, validation, and testing sets.
We ensure each patient’s data belongs to exactly one set to avoid leakage.
We stratify by class to ensure comparable class frequencies.
We repeat this process with three data-split random seeds; each seed selects a different partition into training, validation, and test sets. 

Our \KPSC~data comes from different hospital sites within an integrated healthcare system. 
To better assess cross-site generalization, we assign scans to training, validation, and test sets using site IDs, ensuring each site’s data belongs to exactly one split.
We repeat this data splitting 5 times in a leave-one-site-group-out design.

\textbf{Performance metrics.}
Ultimately, all tables and figures report the test set mean of a specific metric across 3 data splits (5 for \KPSC), with uncertainty quantified via ``+/-" one standard deviation.

All our classification tasks are binary, so we primarily use 
area under the receiver operating characteristic curve (AUROC) to measure discriminative quality.
We further report area under the precision-recall curve (AUPRC) in the supplement. We find that relative method rankings are typically preserved across these metrics.

\textbf{Training and hyperparameter tuning.}
As much as possible, we ensure an apples-to-apples treatment of training and hyperparameter search for fair benchmarking across diverse methods ~\citep{huang2024systematic}.
All methods are trained with mini-batch stochastic gradient descent (SGD) to minimize binary cross entropy loss with L1 or L2 regularization. 
We use SGD with a momentum parameter of 0.9. We set batch size to 64 for frozen MIL and 4 for 3D CNNs (more than 4 risks memory errors on our commodity GPUs).
We train frozen MIL for 1,000 epochs; we train the far more expensive 3D CNNs for 100 epochs, following \citet{liu2019design,liu2022generalizable}.
After training, we select the checkpoint that maximizes validation AUROC, subject to the constraint that validation AUROC is lower than training AUROC, to mitigate unreliable model selection on small datasets.

For both MIL and 3D CNNs, we select learning rate from \{0.1, 0.01, 0.001, 0.0001\} and L1 or L2 regularization strength from \{1.0, 0.1, 0.01, 0.001, 0.0001, 1e-5, 1e-6, 0.0\}.

\section{Results and Analysis}

\setlength{\tabcolsep}{6pt}
\begin{table*}[htbp!]
    \centering
    \caption{Encoder comparison: Test AUROC on OASIS-3 MRI. All MIL methods use \emph{embedding-aggregation}.}
    \label{tab:oasis-3_encoder_auroc}
    \small
    \begin{tabular}{lccccc}
        \hline
        & \multicolumn{3}{c}{\textit{MIL without instance interaction}} & \multicolumn{2}{c}{\textit{MIL with instance interaction}} \\
        & \textbf{Max} & \textbf{Mean} & \textbf{ABMIL} & \textbf{TransMIL} & \textbf{SmAP} \\
        \hline
        ViT-B/16 & $0.802${\tiny$\pm 0.025$} & $0.801${\tiny$\pm 0.035$} & $0.783${\tiny$\pm 0.006$} & $0.744${\tiny$\pm 0.038$} & $0.786${\tiny$\pm 0.022$} \\
        ConvNeXt-Tiny & $0.789${\tiny$\pm 0.039$} & $0.788${\tiny$\pm 0.024$} & $0.801${\tiny$\pm 0.019$} & $0.792${\tiny$\pm 0.032$} & $0.800${\tiny$\pm 0.025$} \\
        MedSAM & $0.769${\tiny$\pm 0.038$} & $0.767${\tiny$\pm 0.003$} & $0.791${\tiny$\pm 0.031$} & $0.775${\tiny$\pm 0.010$} & $0.780${\tiny$\pm 0.026$} \\
        \hline
    \end{tabular}
\end{table*}
\setlength{\tabcolsep}{6pt}

\subsection{Performance on Moderate-Sized Data}
\label{sec:results_medium_data}

Here, we report results and analysis from experiments on moderately-sized CT and MRI datasets (600-3,000 total scans). For such data, thorough experiments were affordable for understanding the impact of different MIL design choices (architectural ordering, encoders, pooling, etc.) as well as tradeoffs between 3D CNNs and MIL methods.
In a later subsection, we analyze the two bigger datasets to see if the best-performing methods still work well.

\begin{figure}[t!]
    \centering
    \includegraphics[width=\linewidth]{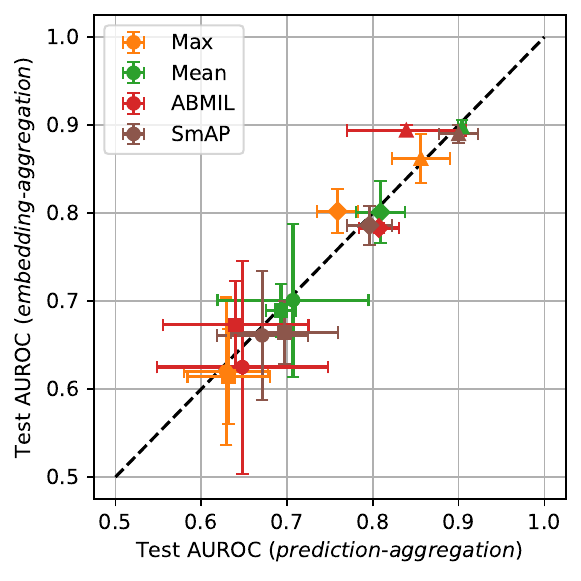}
    \caption{Test AUROC for \emph{embedding-aggregation} and \emph{prediction-aggregation} on OASIS-3 CT (\ding{108}), RSNA-1,149 CT (\ding{115}), ADNI1 MRI (\ding{110}), and OASIS-3 MRI (\ding{117}). \textbf{Takeaway: Results are concentrated around the $y = x$ line, indicating the performance of \emph{embedding-aggregation} and \emph{prediction-aggregation} is comparable across pooling operations and datasets.}}
    \label{fig:architectural_approaches_auroc}
\end{figure}

\textbf{What architectural ordering  works best?}
In Fig.~\ref{fig:architectural_approaches_auroc}, we compare \emph{embedding-aggregation} and \emph{prediction-aggregation} versions of MIL pooling methods, all using a common ViT encoder. We find overall that usually the two orderings yield roughly the same AUROC. Though some individual results can differ by up to 0.05 AUROC, the scale of uncertainty is also high. 
Tables of AUROC and AUPRC numbers for \emph{embedding-aggregation} (Tab.~\ref{tab:medium_datasets_embedding-aggregation_auroc} and \ref{tab:medium_datasets_embedding-aggregation_auprc}) and \emph{prediction-aggregation} (Tab.~\ref{tab:medium_datasets_prediction-aggregation_auroc} and \ref{tab:medium_datasets_prediction-aggregation_auprc}) show that relative ranking of pooling approaches is not greatly altered by the ordering choice.
For the rest of this main paper, we focus on embedding aggregation as it allows inclusion of TransMIL, which does not have a \emph{prediction-aggregation} approach due to the nature of its multi-head self-attention.

\setlength{\tabcolsep}{4pt}
\begin{table*}[htbp!]
    \centering
    \caption{Pooling comparison: Test AUROC on medium datasets. All MIL methods use a frozen ViT and \emph{embedding-aggregation}. \dagger{}3DINO was pre-trained on ADNI1, OASIS-3, and half of the RSNA ICH full dataset.
    }
    \label{tab:medium_datasets_embedding-aggregation_auroc}
    \small
    \resizebox{\linewidth}{!}{\begin{tabular}{ll>{\columncolor{bright_gray}}c>{\columncolor{bright_gray}}ccccc}
        \hline
        & & \multicolumn{2}{c}{\cellcolor{bright_gray} \textbf{CT}} & \multicolumn{4}{c}{\textbf{MRI}} \\
        & & \textbf{OASIS-3} & \textbf{RSNA-1,149} & \textbf{ADNI1} & \textbf{OASIS-3} & \multicolumn{2}{c}{\textbf{\KPSC-800}} \\
        & & AD & ICH & AD & AD & CBI & WMD \\
        \hline
        \multirow{3}{*}{\textit{\makecell[l]{MIL without\\ instance\\interaction}}} & Max & $0.620${\tiny$\pm 0.084$} & $0.862${\tiny$\pm 0.028$} & $0.614${\tiny$\pm 0.054$} & $0.802${\tiny$\pm 0.025$} & $0.648${\tiny$\pm 0.020$} & $0.644${\tiny$\pm 0.052$} \\
        & Mean & $0.701${\tiny$\pm 0.087$} & $0.898${\tiny$\pm 0.008$} & $0.689${\tiny$\pm 0.030$} & $0.801${\tiny$\pm 0.035$} & $0.610${\tiny$\pm 0.050$} & $0.640${\tiny$\pm 0.053$} \\
        & ABMIL & $0.625${\tiny$\pm 0.121$} & $0.894${\tiny$\pm 0.006$} & $0.673${\tiny$\pm 0.050$} & $0.783${\tiny$\pm 0.006$} & $0.609${\tiny$\pm 0.038$} & $0.648${\tiny$\pm 0.020$} \\
        \hline
        \multirow{2}{*}{\textit{\makecell[l]{MIL with\\instance\\interaction}}} & TransMIL & $0.648${\tiny$\pm 0.097$} & $0.893${\tiny$\pm 0.011$} & $0.593${\tiny$\pm 0.067$} & $0.744${\tiny$\pm 0.038$} & $0.565${\tiny$\pm 0.060$} & $0.681${\tiny$\pm 0.054$} \\
        & SmAP & $0.661${\tiny$\pm 0.073$} & $0.890${\tiny$\pm 0.010$} & $0.664${\tiny$\pm 0.036$} & $0.786${\tiny$\pm 0.022$} & $0.610${\tiny$\pm 0.063$} & $0.692${\tiny$\pm 0.077$} \\
        & & & & & & & \\
        \hline
        & 3D ResNet-18 & $0.537${\tiny$\pm 0.079$} & $0.850${\tiny$\pm 0.014$} & $0.567${\tiny$\pm 0.081$} & $0.725${\tiny$\pm 0.029$} & $0.591${\tiny$\pm 0.050$} & $0.563${\tiny$\pm 0.063$} \\
        & 3D CNN for AD & \multicolumn{2}{c}{\cellcolor{bright_gray} \emph{Variable-sized input not supported}} & $0.680${\tiny$\pm 0.043$} & $0.816${\tiny$\pm 0.031$} & $0.689${\tiny$\pm 0.054$} & $0.708${\tiny$\pm 0.077$} \\
        & 3DINO & \dagger{}$0.622${\tiny$\pm 0.038$} & \dagger{}$0.895${\tiny$\pm 0.014$} & \dagger{}$0.581${\tiny$\pm 0.015$} & \dagger{}$0.795${\tiny$\pm 0.051$} & $0.645${\tiny$\pm 0.032$} & $0.691${\tiny$\pm 0.063$} \\
        \hline
    \end{tabular}}
\end{table*}
\setlength{\tabcolsep}{6pt}

\textbf{What encoder works best?}
Tab.~\ref{tab:oasis-3_encoder_auroc} and \ref{tab:oasis-3_encoder_auprc} report AUROC and AUPRC on OASIS-3 MRI for three different encoders: ViT-B/16, ConvNeXt-Tiny, and MedSAM.
Encoder performance varied for each pooling strategy.
ViT-B/16 performed better for simple pooling operations (Max and Mean) while ConvNeXt-Tiny performed better for learnable pooling operations (ABMIL, SmAP, and TransMIL).
MedSAM consistently underperformed both ViT-B/16 and ConvNeXt-Tiny for all pooling operations.
We chose a frozen ViT-B/16 for all subsequent experiments because of its superior performance with simple pooling operations, which was not significantly exceeded by any other encoder with any other pooling strategy.

\textbf{What pooling operation works best?}
Tab.~\ref{tab:medium_datasets_embedding-aggregation_auroc} and \ref{tab:medium_datasets_embedding-aggregation_auprc} report AUROC and AUPRC for \emph{embedding-aggregation} approaches for five different pooling operations: Max, Mean, ABMIL, TransMIL, and SmAP.
On all non-\KPSC\ datasets, Mean pooling performed comparable with and in many cases outperforms, more complicated pooling approaches.
Mean pooling also was competitive in separate evaluations of \emph{prediction-aggregation} approaches (see Tab.~\ref{tab:medium_datasets_prediction-aggregation_auroc} and \ref{tab:medium_datasets_prediction-aggregation_auprc}).
Notably, many prior works on MIL for brain scans~\citep{wu2021combining,lopez2022deep,perez2024end,castro2024sm} do not include Mean pooling as a baseline, despite its strong and consistent performance observed here.

In contrast, for WMD and CBI classification on \KPSC-800, Mean pooling did notably worse than the best 3D CNN and best attention-based MIL.

\textbf{Why does Mean pooling perform well on AD or ICH but not CBI/WMD?}
A possible explanation for the strong performance of Mean pooling is that, for several neuroimaging tasks, signal relevant to classification may be distributed across many slices rather than concentrated in a narrow highly informative region. Therefore, averaging across slices can yield a stable scan-level representation and reduce sensitivity to noise or spurious slice-level artifacts. 

We tentatively hypothesize that Mean pooling performs relatively poorly for CBI and WMD because, unlike the AD label, the imaging signal for these covert cerebrovascular abnormalities can often be focal or regional rather than diffuse.
CBI in particular might appear as a single subcortical lesion that is less than 1cm, particular when due to a lacunar infarct, the most common form of CBI.
Mild WMD (the most common subtype in our data) is often  apparent in  periventricular regions rather than diffusely throughout the brain.
In these settings, diagnostically informative signal is anatomically concentrated, and thus averaging across all slices may dilute signal.
An alternative explanation could be the imperfect nature of the NLP tools used to obtain WMD and CBI labels.

\textbf{How do our reported numbers compare to other published efforts?}
All numbers here were done by our team using our released code. 
It is useful to verify the discriminative performance we report is at least comparable with (if not better than) other efforts published elsewhere.  On RSNA-1,149 dataset, our ABMIL, TransMIL, and SmAP results are comparable to recently published work~\citep{castro2024sm}.
Notably, unlike that work we include the competitive Mean pooling baseline.

As a final sanity check, specifically for the white matter disease (WMD) task on the \KPSC-800 dataset, we compare to SAMSEG~\citep{puonti2016fast,cerri2021contrast,cerri2023open}, a FreeSurfer tool for white matter lesion segmentation.
Using this tool, we can estimate the total volume of white matter as a thresholdable score for WMD classification.
SAMSEG scores $0.699${\tiny$\pm 0.072$} AUROC, within 0.01 of the best results from deep MIL and 3D CNNs in Tab.~\ref{tab:medium_datasets_embedding-aggregation_auroc}.


\subsection{Performance on Large Datasets}
\label{sec:results_large_data}

We now examine the two largest datasets, RSNA-21,744 CT and \KPSC-10k MRI.
Here, the size of the datasets made exhaustive comparisons of all methods with large hyperparameter search grids infeasible.
We thus manually selected an appropriate subset of methods, always using a frozen ViT for MIL.

\setlength{\tabcolsep}{6pt}
\begin{table*}[!ht]
    \centering
    \caption{Pooling comparison: Test AUROC on \textbf{Large} datasets. All MIL methods use a frozen ViT and \emph{embedding-aggregation}.
    Time is for training one NN at one hyperparameter on one train/test split. Results average over 3 data splits for RSNA-21,744 and 5 data splits for \KPSC-10k and show best of a grid search across many hyperparameters.
    }
\label{tab:large_datasets_embedding-aggregation_auroc}
    \small
    \begin{tabular}{ll>{\columncolor{bright_gray}}c>{\columncolor{bright_gray}}lcc}
        \hline
        & & \multicolumn{2}{c}{\cellcolor{bright_gray} \textbf{CT}} & \multicolumn{2}{c}{\textbf{MRI}} \\
        & & \multicolumn{2}{c}{\cellcolor{bright_gray} \textbf{RSNA-21,744}} & \multicolumn{2}{c}{\textbf{\KPSC-10k}} \\
        & & ICH & Time & CBI & WMD \\
        \hline
        \multirow{2}{*}{\textit{\makecell[l]{MIL without\\instance\\interaction}}} 
        & Max 
            & $0.888${\tiny$\pm 0.009$} 
            & \phantom{0 hr. }16 min.
            & $0.644${\tiny$\pm 0.053$} & $0.661${\tiny$\pm 0.020$} \\
        & Mean 
            & $0.920${\tiny$\pm 0.012$} 
            & \phantom{0 hr. }21 min.
            & $0.666${\tiny$\pm 0.017$} & $0.689${\tiny$\pm 0.027$} \\
        & ABMIL 
            & $0.919${\tiny$\pm 0.009$} 
            & \phantom{0 hr. }42 min.
            & $0.647${\tiny$\pm 0.055$} 
            & $0.713${\tiny$\pm 0.033$} \\
        \hline
        \multirow{2}{*}{\textit{\makecell[l]{MIL with\\instance\\interaction}}} 
        & TransMIL 
            & $0.925${\tiny$\pm 0.014$} 
            & 9 hr. 13 min.
            & $0.637${\tiny$\pm 0.056$} & $0.715${\tiny$\pm 0.030$} \\
        & SmAP 
            & $0.925${\tiny$\pm 0.012$} 
            & 8 hr. 43 min.
            & $0.669${\tiny$\pm 0.050$} & $0.714${\tiny$\pm 0.031$} \\
        & & & & & \\
        \hline
        & 3D CNN for AD & 
            \multicolumn{2}{c}{\cellcolor{bright_gray} \emph{Variable-sized input not supported}} 
            & $0.675${\tiny$\pm 0.074$} 
            & $0.724${\tiny$\pm 0.030$} \\
        \hline
    \end{tabular}
\end{table*}
\setlength{\tabcolsep}{6pt}

\textbf{What works on big CT?}
Here, we focused on deep MIL comparisons.
The \emph{3D CNN for AD} cannot be used out-of-the-box on RSNA-21,744 because each CT scan has a variable number of slices.
The 3D ResNet-18 performed poorly on moderately-sized CT data, so we elected to skip it here.

Tab.~\ref{tab:large_datasets_embedding-aggregation_auroc} shows that the core message of our Fig.~\ref{fig:oasis-3_mri_number_of_trainable_parameters} is true even with 10,000+ scans: more flexible MIL methods like TransMIL can outperform simplistic MIL, but only by a modest margin (less than 0.01 AUROC gain on CT) while requiring 25x the compute. 
For many practitioners, it may be question whether such modest gains are worth the effort over just using the strong 0.922 AUROC from \emph{Mean pooling MIL}.

\textbf{What works on big MRI?}
The \emph{3D CNN for AD} was competitive on moderately-sized MRI datasets, so we compared it to MIL on the Proprietary-10k MRI dataset.
Tab.~\ref{tab:large_datasets_embedding-aggregation_auroc} shows that 3D CNNs can outperform attention-based MIL, but they are more expensive to train.
Notably, Mean pooling is competitive and much more efficient to train.

\subsection{Evaluating Per-Slice Attention Quality}
\label{sec:results_attention_quality}

Learnable attention is what differentiates the advanced MIL methods of recent years from simplistic Mean pooling MIL. 
Here, we take advantage of a unique aspect of the large RSNA dataset: every scan has instance-level labels indicating the positive/negative status of each slice of the 3D scan for the ICH binary task.
While the attention value at a slice is not necessarily intended to exactly mean the predicted positive-class  probability in any MIL network, the basic logic of MIL prediction suggests that when a 3D scan is positive, at least some attention should be paid to positive slices.
Past works have thus assessed per-instance attention as a predictor of instance-level class label \citep{castro2024sm}.

\textbf{Metrics.}
To study how well attention values $a_{ij}$ produced by MIL may predict the binary instance-level ICH labels $y_{ij}$, we report three higher-is-better metrics: (1)~attention correctness \citep{liu2017attention}, (2) AUROC, and (3)~AUPRC.

\textbf{Center-focused baseline.}
Beyond the MIL methods, we compare to a simple baseline that always allocates attention via a Gaussian bell-shaped curve centered at the middle slice of the 3D scan, regardless of the input image.
This is meant to provide a ``center-focused'' inductive bias, as a brief exploratory analysis of RSNA per-slice labels suggests positive instances often occur close to the middle of the axial scan, reflecting the typical neuroanatomic distribution of ICH lesions.

\textbf{Result: MIL attention quality does not exceed simple baseline.}
Results are found in Tab.~\ref{tab:rsna_instance-level_metrics}.
Among attention-based MIL methods, we find that TransMIL and SmAP substantially outperform ABMIL on instance-level AUROC and AUPRC.
However, the center-focused baseline, which does not depend on the input neuroimage at all, surprisingly achieves the best results overall, exceeding the next-best method by $0.154$ in attention correctness and $0.019$ AUROC.

\setlength{\tabcolsep}{6pt}
  \begin{table*}[htbp!]
      \centering
      \caption{Per-slice attention quality metrics on test set of RSNA-21,744. All MIL methods use a frozen ViT and \emph{embedding-aggregation}.}
      \label{tab:rsna_instance-level_metrics}
      \small
      \begin{tabular}{l>{\columncolor{bright_gray}}cccccc}
          \hline
          & \textbf{Centered Gaussian} 
          & \textbf{Max} & \textbf{Mean} &  \textbf{ABMIL} & \textbf{TransMIL} & \textbf{SmAP} \\
          \hline
          Attention correctness 
            & $0.701${\tiny$\pm 0.011$} 
            & N/A 
            & $0.359${\tiny$\pm 0.009$} 
            & $0.547${\tiny$\pm 0.087$} 
            & $0.435${\tiny$\pm 0.028$} 
            & $0.509${\tiny$\pm 0.010$} \\
          AUROC 
            & $0.850${\tiny$\pm 0.001$} 
            & N/A 
            & $0.500${\tiny$\pm 0.000$} 
            & $0.736${\tiny$\pm 0.052$} 
            & $0.792${\tiny$\pm 0.031$} 
            & $0.831${\tiny$\pm 0.003$} \\
          AUPRC 
            & $0.710${\tiny$\pm 0.007$}
            & N/A 
            & $0.359${\tiny$\pm 0.009$} 
            & $0.634${\tiny$\pm 0.049$} 
            & $0.659${\tiny$\pm 0.032$} 
            & $0.713${\tiny$\pm 0.007$} \\
          \hline
      \end{tabular}
  \end{table*}
  \setlength{\tabcolsep}{6pt}

We use bootstrapping \citep{foody2009classification} to access the statistical significance of some AUROC difference.

\section{Semi-Synthetic Experiments}
\label{sec:semi_synth}

\begin{figure*}[htbp!]
    \centering
    \includegraphics[width=\linewidth]{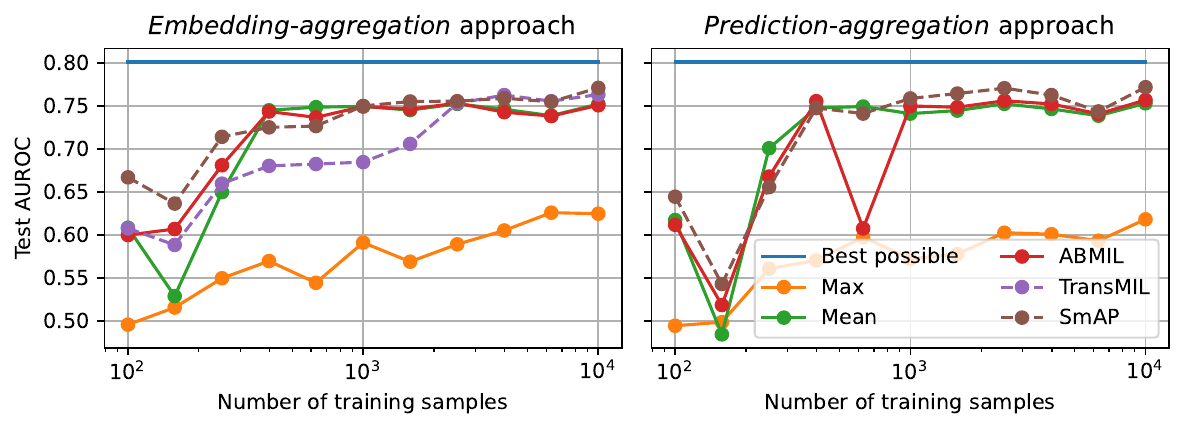}
    \caption{Test AUROC vs.~train set size for semi-synthetic Shifted Mean task (Sec.~\ref{sec:semi_synth}), where aggregating subtle signals across 12  instances is needed. \textbf{Takeaway: Even with 10,000 training scans,  SmAP and TransMIL barely outperform Mean pooling and fall at least 0.04 AUROC below the best possible Bayes estimator.}
    }
    \label{fig:shifted_mean_mil_varying_number_of_training_samples}
\end{figure*}

To better understand the effectiveness of mean pooling, we design a semi-synthetic dataset and compare MIL variants on it. We intend this data to represent key challenges in MIL for 3D brain scans: 
(1)~only some features in an embedding are discriminative, (2)~only a few instances in a bag signal whether it should have a positive label, and (3)~context from nearby instances matters, as the information from an individual instance may be statistically ambiguous.
We set key data statistics to match the RSNA dataset: we match the range of the number of instances in a bag: $S_i \in \{20, \dots 60\}$, set the mean number of contiguous positive instances to $R = 12$ as in RSNA, and use 768 features like a ViT embedding.

This effort is similar in spirit to  work on \emph{algorithmic unit tests} for MIL~\citep{raff2023reproducibility}.
That paper uses synthetic classification tasks designed to reveal whether learned models violate key MIL assumptions, such as a bag is positive if and only if one or more instances have a positive label.
Our new data focuses instead on assessing context from nearby instances and quantifying best possible performance.

\textbf{Data generation.} 
We define a data-generating process we call \emph{Shifted Mean MIL} that jointly samples $h_{i, 1:S_i}, y_i$ embedding-label pairs. 
We draw $y_i \sim \operation{Bern}(0.5)$, so 50\% of scans are positive, then draw the number of instances $S_i \sim \operation{Unif}(\{20, \dots, 60\})$. If the scan is negative, all embeddings for all instances are drawn from a mean=0, variance=1 Gaussian.
If positive, we select a contiguous block of $R = 12$ instances to indicate positivity, and draw their first feature (of many) from a Gaussian with mean $\Delta = 0.5$; other features are drawn from a zero-mean Gaussian.  
We intentionally set $\Delta$ low so that no one instance will be a ``smoking gun,'' but reasoning over $R$ instances should reliably indicate which bags are positives.
See App.~\ref{sec:shifted_mean_mil_dataset} for dataset details.

Given the true distribution, we characterize the best possible classifier for this data by deriving a Bayes estimator \citep{degroot1970optimal,murphy2022BayesianDecisionTheory}, the best probabilistic predictor of $y_i$ given $h_i$ for this data.
See App.~\ref{sec:bayes_estimator} for details.

\textbf{Experiments.}
We intend the provided $h_i$ already represent encoder-provided embeddings. Thus, we assess how well MIL pooling and classification variants can classify as a function of the total number of training samples, which we step from 100 to 10,000.
At each size, we randomly assign bags using an 4:1 ratio into training and validation sets. 
We report results on a fixed separate test set of 1,000 bags.

\textbf{Results: Can current MIL methods capture instance interaction?}
Across training sample sizes in Fig.~\ref{fig:shifted_mean_mil_varying_number_of_training_samples}, even modern interaction-aware MIL approaches may fail to match Bayes-optimal performance. 
Neither TransMIL nor SmAP offers any noticeable gains over Mean-pooling, and all fall well short (at least 0.05 AUROC below) the best possible blue line of the Bayes estimator. 
Capturing subtle contextual dependence across instances appears challenging for current off-the-shelf MIL even at large data sizes. 
This helps explain why mean pooling works well for 3D brain classification tasks that have subtle signals at individual slices yet many instances (a dozen or more) that could indicate positivity.

\section{Conclusion}
\label{sec:conclusion}

We presented a benchmark of simple MIL, attention-based MIL, and 3D CNNs on 9 classification tasks across 7 datasets.
Our results highlight that MIL with Mean pooling is often a strong baseline that takes less than an hour to train even on the largest dataset we tested (21,744 scans), while the latest attention-based models require 8x or more training time per run to deliver modest gains of roughly +0.01-0.03 to the AUROC value.
Instead of immediately reaching for bigger models,
practitioners may explore what else they could do with those hours, such as pursue embedding-based augmentation strategies~\citep{verma2019manifold}, consider deep ensembles~\citep{lakshminarayanan2017simple} that average over encoders, or seek additional data sources.

Another key message of our work, supported by the per-slice attention quality experiments in Sec.~\ref{sec:results_attention_quality}, is that at least for 3D brain scans, the learned attention of modern well-published methods like TransMIL or SmAP does not seem to find reliable per-image signals. Future work might seek to inform attention-based MIL via the inductive bias of our Centered Gaussian baseline.
Our semi-synthetic results in Sec.~\ref{sec:semi_synth} further suggest that, especially for brain tasks where careful examination of many relevant slices is necessary, current MIL scores substantially below what is possible even with transformer-based attention and 10,000 training scans. Improved inductive biases and regularization strategies are needed.

Our work has many limitations. 
All analyses use recommended data preprocessing steps and did not consider alternatives. 
Our results mostly focus on  encoders pretrained on non-brain images without further fine-tuning; we found fine-tuning was not worth the effort (see Tab.~\ref{tab:oasis-3_encoder_auroc}) but others may wish to revisit this.
We offer comparison to a few well-published 3D CNNs, but our list of 3D CNN methods is not exhaustive and others may perform better.

Ultimately, practitioners in the future may rely more on pretrained foundation models for 3D brain scan classification. 
We hope our work provides a path forward for resource-constrained analysts to quickly reach competitive prediction quality and for method developers to improve per-slice attention mechanisms and overall pooling strategies.

\section*{Acknowledgments}
This work is supported by the U.S. National Institutes of Health (grant \# R01NS134859) and the Alzheimer’s Drug Discovery Foundation. Author MCH is also supported in part by the U.S. National Science Foundation (NSF) via IIS CAREER grant \# 2338962. 
Author AAS is supported by an NIH postdoctoral T32 award (T32TR004418).
We are grateful for resources and support from the Tufts High-Performance Computing Cluster.
This paper's content is solely the responsibility of the authors and does not necessarily represent the official views of the NIH or NSF.

We are thankful for computing infrastructure support provided by Research Technology Services at Tufts University, with hardware funded in part by NSF award OAC CC* \# 2018149.

\newpage

\bibliography{main}

\onecolumn
\appendix

\counterwithin{table}{section}
\setcounter{table}{0}
\counterwithin{figure}{section}
\setcounter{figure}{0}

\section{Preprocessing}
\label{sec:preprocessing}

\subsection{CT}

For CT images, we convert images into Hounsfield Units (HU) using each image's rescale slope and intercept; skull strip images (only including -100 to 300 HU) \citep{muschelli2019recommendations}; resize each 2D slice to $224 \times 224$ pixels for VIT-B/16 and ConvNeXt-Tiny, and $1024 \times 1024$ pixels for MedSAM; and normalize images with the training set mean and standard deviation of each channel.

\subsection{MRI}

For MRI images, we follow the \texttt{t1-linear} pipeline from Clinica \citep{wen2020convolutional,routier2021clinica}.
We correct bias field inhomogeneities using the N4ITK method \citep{tustison2010n4itk}; register each image to the MNI space with the ICBM 2009c nonlinear symmetric template \citep{fonov2009unbiased,fonov2011unbiased} using the SyN algorithm \citep{avants2008symmetric} from ANTs \citep{avants2014insight}; crop each image to remove background; resize each 2D slice to $224 \times 224$ pixels for VIT-B/16 and ConvNeXt-Tiny, and $1024 \times 1024$ pixels for MedSAM; and normalize images with the training set mean and standard deviation of each channel.

\section{\emph{Prediction-Aggregation} AUROC Results}
\label{sec:prediction-aggregation_approach_auroc_results}

\setlength{\tabcolsep}{6pt}
\begin{table*}[htbp!]
    \centering
    \caption{Test AUROC on moderately-sized datasets. All MIL methods use the \emph{prediction-aggregation} approach.}
    \label{tab:medium_datasets_prediction-aggregation_auroc}
    \small
    \begin{tabular}{ll>{\columncolor{bright_gray}}c>{\columncolor{bright_gray}}ccccc}
        \hline
        & & \multicolumn{2}{c}{\cellcolor{bright_gray} \textbf{CT}} & \multicolumn{4}{c}{\textbf{MRI}} \\
        & & \textbf{OASIS-3} & \textbf{RSNA Subset} & \textbf{ADNI1} & \textbf{OASIS-3} & \multicolumn{2}{c}{\textbf{\KPSC-800}} \\
        & & AD & ICH & AD & AD & CBI & WMD \\
        \hline
        \multirow{3}{*}{\textit{\makecell[l]{MIL without\\instance\\interaction}}} & Max & $0.629${\tiny$\pm 0.049$} & $0.856${\tiny$\pm 0.034$} & $0.632${\tiny$\pm 0.048$} & $0.759${\tiny$\pm 0.024$} & $0.553${\tiny$\pm 0.063$} & $0.642${\tiny$\pm 0.059$} \\
        & Mean & $0.707${\tiny$\pm 0.088$} & $0.904${\tiny$\pm 0.005$} & $0.693${\tiny$\pm 0.017$} & $0.809${\tiny$\pm 0.029$} & $0.620${\tiny$\pm 0.034$} & $0.627${\tiny$\pm 0.070$} \\
        & ABMIL & $0.648${\tiny$\pm 0.100$} & $0.839${\tiny$\pm 0.069$} & $0.640${\tiny$\pm 0.085$} & $0.807${\tiny$\pm 0.023$} & $0.599${\tiny$\pm 0.038$} & $0.664${\tiny$\pm 0.066$} \\
        \hline
        \multirow{3}{*}{\textit{\makecell[l]{MIL with\\instance\\interaction}}} & SmAP & $0.671${\tiny$\pm 0.053$} & $0.900${\tiny$\pm 0.023$} & $0.697${\tiny$\pm 0.062$} & $0.796${\tiny$\pm 0.026$} & $0.602${\tiny$\pm 0.016$} & $0.664${\tiny$\pm 0.063$} \\
        & & & & & & & \\
        & & & & & & & \\
        \hline
    \end{tabular}
\end{table*}
\setlength{\tabcolsep}{6pt}

\section{AUPRC Results}
\label{sec:auprc_results}

\setlength{\tabcolsep}{6pt}
\begin{table*}[htbp!]
    \centering
    \caption{Test AUPRC on OASIS-3 MRI. All MIL methods use the \emph{embedding-aggregation} approach.}
    \label{tab:oasis-3_encoder_auprc}
    \small
    \begin{tabular}{lccccc}
        \hline
        & \multicolumn{3}{c}{\textit{MIL without instance interaction}} & \multicolumn{2}{c}{\textit{MIL with instance interaction}} \\
        & \textbf{Max} & \textbf{Mean} & \textbf{ABMIL} & \textbf{TransMIL} & \textbf{SmAP} \\
        \hline
        ViT-B/16 & $0.562${\tiny$\pm 0.081$} & $0.613${\tiny$\pm 0.079$} & $0.555${\tiny$\pm 0.028$} & $0.498${\tiny$\pm 0.085$} & $0.570${\tiny$\pm 0.065$} \\
        ConvNeXt-Tiny & $0.562${\tiny$\pm 0.086$} & $0.531${\tiny$\pm 0.037$} & $0.531${\tiny$\pm 0.037$} & $0.571${\tiny$\pm 0.115$} & $0.579${\tiny$\pm 0.064$} \\
        MedSAM & $0.509${\tiny$\pm 0.102$} & $0.510${\tiny$\pm 0.013$} & $0.547${\tiny$\pm 0.007$} & $0.522${\tiny$\pm 0.027$} & $0.515${\tiny$\pm 0.044$} \\
        \hline
    \end{tabular}
\end{table*}
\setlength{\tabcolsep}{6pt}

\setlength{\tabcolsep}{4pt}
\begin{table*}[htbp!]
    \centering
    \caption{Test AUPRC moderately-sized datasets. All MIL methods use the \emph{embedding-aggregation} approach. \dagger{}3DINO was pre-trained on ADNI1, OASIS-3, and half of the RSNA ICH full dataset.}
    \label{tab:medium_datasets_embedding-aggregation_auprc}
    \small
    \resizebox{\linewidth}{!}{\begin{tabular}{ll>{\columncolor{bright_gray}}c>{\columncolor{bright_gray}}ccccc}
        \hline
        & & \multicolumn{2}{c}{\cellcolor{bright_gray} \textbf{CT}} & \multicolumn{4}{c}{\textbf{MRI}} \\
        & & \textbf{OASIS-3} & \textbf{RSNA Subset} & \textbf{ADNI1} & \textbf{OASIS-3} & \multicolumn{2}{c}{\textbf{\KPSC-800}} \\
        & & AD & ICH & AD & AD & CBI & WMD \\
        \hline
        \multirow{3}{*}{\textit{\makecell[l]{MIL without\\instance\\interaction}}} & Max & $0.303${\tiny$\pm 0.095$} & $0.821${\tiny$\pm 0.032$} & $0.315${\tiny$\pm 0.066$} & $0.562${\tiny$\pm 0.081$} & $0.452${\tiny$\pm 0.111$} & $0.769${\tiny$\pm 0.114$} \\
        & Mean & $0.347${\tiny$\pm 0.078$} & $0.872${\tiny$\pm 0.008$} & $0.326${\tiny$\pm 0.038$} & $0.613${\tiny$\pm 0.079$} & $0.413${\tiny$\pm 0.108$} & $0.776${\tiny$\pm 0.097$} \\
        & ABMIL & $0.305${\tiny$\pm 0.057$} & $0.874${\tiny$\pm 0.015$} & $0.325${\tiny$\pm 0.030$} & $0.555${\tiny$\pm 0.028$} & $0.412${\tiny$\pm 0.113$} & $0.787${\tiny$\pm 0.070$} \\
        \hline
        \multirow{3}{*}{\textit{\makecell[l]{MIL with\\instance\\interaction}}} & TransMIL & $0.292${\tiny$\pm 0.061$} & $0.866${\tiny$\pm 0.001$} & $0.260${\tiny$\pm 0.025$} & $0.498${\tiny$\pm 0.085$} & $0.392${\tiny$\pm 0.122$} & $0.807${\tiny$\pm 0.090$} \\
        & SmAP & $0.356${\tiny$\pm 0.029$} & $0.867${\tiny$\pm 0.029$} & $0.323${\tiny$\pm 0.062$} & $0.570${\tiny$\pm 0.065$} & $0.422${\tiny$\pm 0.114$} & $0.809${\tiny$\pm 0.092$} \\
        & & & & & & & \\
        \hline
        & 3D ResNet-18 & $0.233${\tiny$\pm 0.051$} & $0.789${\tiny$\pm 0.037$} & $0.245${\tiny$\pm 0.053$} & $0.439${\tiny$\pm 0.027$} & $0.393${\tiny$\pm 0.087$} & $0.729${\tiny$\pm 0.104$} \\
        & 3D CNN for AD & \multicolumn{2}{c}{\cellcolor{bright_gray} \emph{Variable-sized input not supported}} & $0.303${\tiny$\pm 0.016$} & $0.636${\tiny$\pm 0.016$} & $0.520${\tiny$\pm 0.133$} & $0.817${\tiny$\pm 0.114$} \\
        & 3DINO & \dagger{}$0.283${\tiny$\pm 0.039$} & \dagger{}$0.887${\tiny$\pm 0.015$} & \dagger{}$0.284${\tiny$\pm 0.032$} & \dagger{}$0.530${\tiny$\pm 0.103$} & $0.597${\tiny$\pm 0.027$} & $0.693${\tiny$\pm 0.061$} \\
        \hline
    \end{tabular}}
\end{table*}
\setlength{\tabcolsep}{6pt}

\setlength{\tabcolsep}{6pt}
\begin{table*}[htbp!]
    \centering
    \caption{Test AUPRC on moderately-sized datasets. All MIL methods use the \emph{prediction-aggregation} approach.}
    \label{tab:medium_datasets_prediction-aggregation_auprc}
    \small
    \begin{tabular}{ll>{\columncolor{bright_gray}}c>{\columncolor{bright_gray}}ccccc}
        \hline
        & & \multicolumn{2}{c}{\cellcolor{bright_gray} \textbf{CT}} & \multicolumn{4}{c}{\textbf{MRI}} \\
        & & \textbf{OASIS-3} & \textbf{RSNA Subset} & \textbf{ADNI1} & \textbf{OASIS-3} & \multicolumn{2}{c}{\textbf{\KPSC-800}} \\
        & & AD & ICH & AD & AD & CBI & WMD \\
        \hline
        \multirow{3}{*}{\textit{\makecell[l]{MIL without\\instance\\interaction}}} & Max & $0.321${\tiny$\pm 0.134$} & $0.823${\tiny$\pm 0.029$} & $0.290${\tiny$\pm 0.039$} & $0.511${\tiny$\pm 0.082$} & $0.392${\tiny $\pm 0.103$} & $0.784${\tiny $\pm 0.102$} \\
        & Mean & $0.328${\tiny$\pm 0.077$} & $0.876${\tiny$\pm 0.008$} & $0.334${\tiny$\pm 0.043$} & $0.618${\tiny$\pm 0.078$} & $0.418${\tiny $\pm 0.108$} & $0.761${\tiny $\pm 0.112$} \\
        & ABMIL & $0.348${\tiny$\pm 0.038$} & $0.819${\tiny$\pm 0.076$} & $0.322${\tiny$\pm 0.079$} & $0.596${\tiny$\pm 0.046$} & $0.406${\tiny $\pm 0.104$} & $0.791${\tiny $\pm 0.095$} \\
        \hline
        \multirow{3}{*}{\textit{\makecell[l]{MIL with\\instance\\interaction}}} & SmAP & $0.330${\tiny$\pm 0.029$} & $0.888${\tiny$\pm 0.024$} & $0.366${\tiny$\pm 0.023$} & $0.579${\tiny$\pm 0.033$} & $0.404${\tiny$\pm 0.098$} & $0.788${\tiny$\pm 0.099$} \\
        & & & & & & & \\
        & & & & & & & \\
        \hline
    \end{tabular}
\end{table*}
\setlength{\tabcolsep}{6pt}

\setlength{\tabcolsep}{6pt}
\begin{table*}[htbp!]
    \centering
    \caption{Test AUPRC on large datasets. All MIL methods use the \emph{embedding-aggregation} approach.}
    \label{tab:large_datasets_embedding-aggregation_auprc}
    \small
    \begin{tabular}{ll>{\columncolor{bright_gray}}ccc}
        \hline
        & & \textbf{CT} & \multicolumn{2}{c}{\textbf{MRI}} \\
        & & \textbf{RSNA-21,744} & \multicolumn{2}{c}{\textbf{\KPSC-10k}} \\
        & & ICH & CBI & WMD \\
        \hline
        \multirow{3}{*}{\textit{\makecell[l]{MIL without\\instance\\interaction}}} & Max & $0.859${\tiny$\pm 0.013$} & $0.073${\tiny$\pm 0.020$} & $0.711${\tiny$\pm 0.068$} \\
        & Mean & $0.903${\tiny$\pm 0.011$} & $0.093${\tiny$\pm 0.022$} & $0.741${\tiny$\pm 0.059$} \\
        & ABMIL & $0.903${\tiny$\pm 0.007$} & $0.078${\tiny$\pm 0.023$} & $0.772${\tiny$\pm 0.051$} \\
        \hline
        \multirow{3}{*}{\textit{\makecell[l]{MIL with\\instance\\interaction}}} & TransMIL & $0.912${\tiny$\pm 0.012$} & $0.088${\tiny$\pm 0.028$} & $0.773${\tiny$\pm 0.052$} \\
        & SmAP & $0.910${\tiny$\pm 0.011$} & $0.102${\tiny$\pm 0.039$} & $0.773${\tiny$\pm 0.054$} \\
        & & & & \\
        \hline
        & 3D CNN for AD & \emph{Variable-sized input not supported} & $0.113${\tiny$\pm 0.035$} & $0.769${\tiny$\pm 0.041$} \\
        \hline
    \end{tabular}
\end{table*}
\setlength{\tabcolsep}{6pt}

\newpage
\section{Instance-Level Results}
\label{sec:attention_visualizations}

\subsection{Attention Quality Metric Definitions}

For each positive bag $i$ ($y_i = 1$) with $S_i$ instances, let $y_{ij} \in \{0,1\}$ denote the instance-level label and $a_{ij}$ the attention weight for instance $j$.
We compute three metrics using attention weights as predictors of instance labels: (1) \emph{attention correctness}~\citep{liu2017attention} $\sum_{j=1}^{S_i} a_{ij} \cdot y_{ij}$, (2) AUROC treating $a_{ij}$ as scores for predicting $y_{ij}$, and (3) AUPRC similarly.
Each metric is computed per positive bag, then averaged across all positive bags in the test set.

\subsection{Attention Visualizations}

\begin{figure}[htbp!]
    \centering
    \includegraphics[width=0.4968\linewidth]{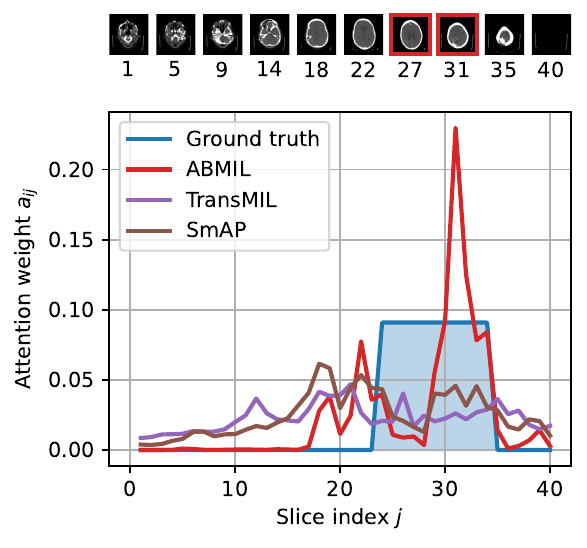}
    \includegraphics[width=0.4968\linewidth]{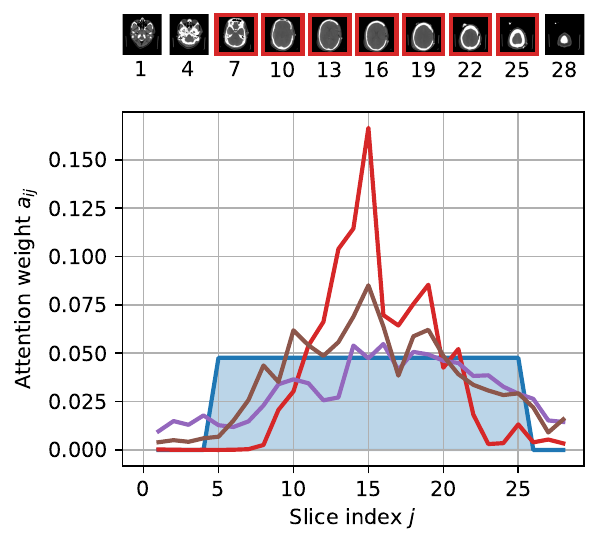}
    \includegraphics[width=0.4968\linewidth]{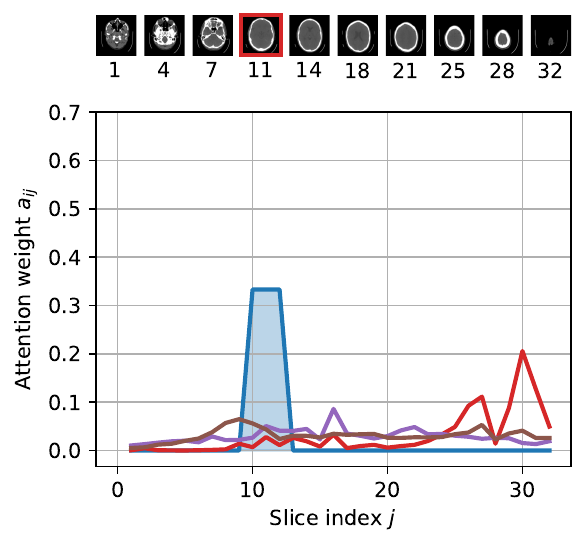}
    \includegraphics[width=0.4968\linewidth]{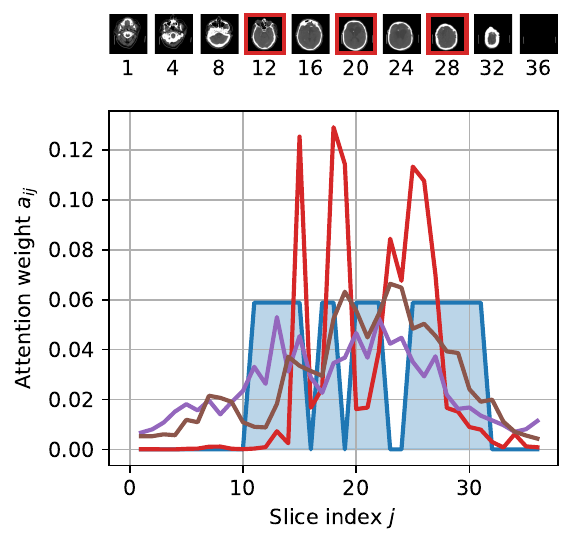}
    \caption{Slice-level attention analysis on test scans. The ground truth has attention uniformly distributed on positive slices and the positive slice images are outlined in red.}
    \label{fig:attention_visualizations}
\end{figure}

\newpage
\section{Shifted Mean MIL Dataset}
\label{sec:shifted_mean_mil_dataset}

\begin{figure*}[htbp!]
    \centering
    \includegraphics[width=\linewidth]{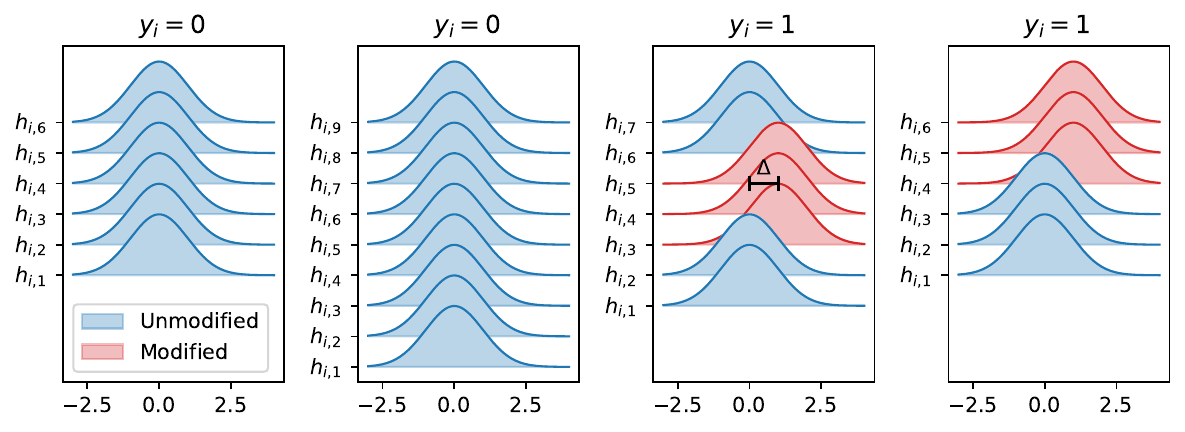}
    \caption{Example data-generating distributions for a discriminative feature for negative ($y_i = 0$) and positive ($y_i = 1)$ bags of $S_i$ instances drawn from our Shifted Mean MIL dataset.
    }
    \label{fig:synthetic_data_visualization}
\end{figure*}

We propose a new data-generating process designed to mimic several key challenges in real-world multiple instance medical imaging tasks:
\begin{itemize}
    \item Across the whole dataset, only some features are discriminative ($K$ of $M$).
    \item For each positively-labeled bag, only a few instances are relevant ($R$ of $S_i$) and they are adjacent in a known 1D listing of all $S_i$ instances.
    \item Context matters. Adjacent instances together provide stronger statistical signal than any one relevant instance's discriminative feature value alone.
\end{itemize}
The generative process for bag $i$ first draws the bag's binary label and the number of instances in the bag
\begin{align}
    y_i &\sim \operation{Bern}(q_{+}), \quad S_i \sim \operation{Unif}(\{S_{\text{low}}, \dots, S_{\text{high}}\}).
\end{align}
Next, for negative bags we sample all features $k$ for all instances $j$ independently from a common Gaussian:
\begin{align}
    h_{ijk} \mid y_i {=} 0 \sim \mathcal{N}( \mu, \sigma^2).
\end{align}
For positive bags, most instances and features are sampled from this same Gaussian. However, for the $K$ discriminative features, we select $R$ adjacent instances (using $u_i$ to denote the starting index) and sample these from a Gaussian with \emph{shifted mean}:
\begin{align}
    \label{eq:u_given_y}
    u_i \mid S_i, y_i{=}1 &\sim \operation{Unif}(\{1, \dots, S_i{-}R{+}1\}),
    \\
    \label{eq:h_given_u_y}
    h_{ijk} \mid u_i, y_i {=} 1
    & \sim
    \begin{cases}
        \mathcal{N}(\mu + \Delta, \sigma^2), 
        & \text{if $j \in [u_i, u_i{+}R{-}1]$ and $k$ is discriminative}
        \\
        \mathcal{N}(\mu, \sigma^2), & \text{otherwise}.
    \end{cases}
\end{align}
Here $\Delta > 0$ indicates the magnitude of shift for discriminative features. Setting $R > 1$ indicates that context helps.
Given a fixed $\mu$, bags drawn from this process are more challenging to classify (even with knowledge of the true process) when $\Delta$ is smaller, $R$ is smaller, $\frac{K}{M}$ is smaller, and $\sigma$ is larger.

This data-generating process is illustrated in Fig.~\ref{fig:synthetic_data_visualization}, depicting only one feature that is discriminative.
In each positive bag, a different contiguous block of $R = 3$ instances draw from the shifted mean Gaussian.
If future work wanted to model correlations between features within an instance, the sampling of vector $h_{ij}$ in Eq.~\eqref{eq:h_given_u_y} could be modified to draw from a multivariate Gaussian with a non-diagonal covariance matrix.

\section{Bayes Estimator}
\label{sec:bayes_estimator}

Given a data-generating process, a \emph{Bayes estimator} is a decision rule that minimizes the posterior expected loss with respect to the data-generating distribution \citep{degroot1970optimal,murphy2022BayesianDecisionTheory}.
It is an oracle upper bound on performance.
By comparing conventional or recent MIL methods to the Bayes estimator for our synthetic dataset, we can quantify how close they come to the best possible performance.

Given a bag $h_i$ of $S_i$ instances and assuming our data-generating process defined above, the Bayes estimator of the posterior probability is:
\begin{align}
    p(y_i{=}1 \mid h_i, S_i) = \frac{p(h_i \mid S_i, y_i{=}1) p(S_i) p(y_i{=}1)}{p(h_i \mid S_i, y_i{=}0) p(S_i) p(y_i{=}0) + p(h_i \mid S_i, y_i{=}1) p(S_i) p(y_i{=}1)}.
\end{align}
Each term on the right-hand side can be computed in closed-form.
The class-conditional likelihood for the negative class factors over instances:
\begin{align}
    p(h_i \mid S_i, y_i{=}0) = \prod_{j=1}^{S_i} \prod_{k=1}^{M} \mathcal{N}(h_{ijk} \mid \mu, \sigma^2).
\end{align}
Each positive bag has a latent segment of $R$ consecutive relevant instances. The class-conditional likelihood for the positive class marginalizes out the unknown index $u$:
\begin{align}
    p(h_i \mid S_i, y_i{=}1) = \sum_{u=1}^{S_i-R+1} \left[ 
    p(u \mid S_i, y_i{=}1)
    \prod_{j=1}^{S_i} \prod_{k=1}^{M} p(h_{ijk} \mid u, y_i{=}1) \right] .
\end{align}
Eq.~\eqref{eq:u_given_y} and \eqref{eq:h_given_u_y} provide the necessary PDF values to evaluate the right hand side.

\end{document}